\documentclass[electronics,article,accept,pdftex,moreauthors]{Definitions/mdpi} 

\usepackage{balance}
\usepackage{amsmath,amssymb,amsfonts}
\usepackage{algorithmic}
\usepackage{graphicx}
\usepackage{textcomp}
\usepackage{xcolor}
\usepackage{amssymb}
\usepackage{pifont}
\newcommand{\cmark}{\ding{51}}
\newcommand{\xmark}{\ding{55}}
\firstpage{1} 
\pubvolume{1}
\issuenum{1}
\articlenumber{0}
\pubyear{2025}
\copyrightyear{2025}
\datereceived{} 
\daterevised{} 
\dateaccepted{ } 
\datepublished{ } 
\hreflink{https://doi.org/} 

\Title{Vision Transformers for Efficient Indoor Pathloss Radio Map Prediction}

\TitleCitation{Vision Transformers for Efficient Indoor Pathloss Radio Map Prediction}

\Author{Rafayel Mkrtchyan $^{1,2}$\orcidC{}, Edvard Ghukasyan $^{1,2}$, Khoren Petrosyan $^{1,2}$, Hrant Khachatrian $^{1,2}$\orcidB{} \mbox{and Theofanis P. Raptis $^{3,}$*\orcidA{}}}

\AuthorNames{Rafayel Mkrtchyan, Edvard Ghukasyan, Khoren Petrosyan, Hrant Khachatrian and Theofanis P. Raptis}

\isAPAStyle{%
       \AuthorCitation{Lastname, F., Lastname, F., \& Lastname, F.}
         }{%
        \isChicagoStyle{%
        \AuthorCitation{Lastname, Firstname, Firstname Lastname, and Firstname Lastname.}
        }{
        \AuthorCitation{Mkrtchyan, R.; Ghukasyan, E.; Petrosyan, K.; Khachatrian, H.; Raptis, T.P.}
        }
}

\address{%
$^{1}$ \quad Machine Learning Group, Center for Mathematical and Applied Research, Yerevan State University,
Yerevan 0025, Armenia
\\
$^{2}$ \quad YerevaNN,  Yerevan 0025,  Armenia\\
$^{3}$ \quad Institute of Informatics and Telematics, National Research Council, 56124 Pisa,  Italy}

\corres{Correspondence: theofanis.raptis@iit.cnr.it}

\abstract{Indoor pathloss prediction is a fundamental task in wireless network planning, yet it remains challenging due to environmental complexity and data scarcity. In this work, we propose a deep learning-based approach utilizing a vision transformer (ViT) architecture with DINO-v2 pretrained weights to model indoor radio propagation. Our method processes a floor map with additional features of the walls 
to generate indoor pathloss maps. We systematically evaluate the effects of architectural choices, data augmentation strategies, and feature engineering techniques. Our findings indicate that extensive augmentation significantly improves generalization, while feature engineering is crucial in low-data regimes. Through comprehensive experiments, we demonstrate the robustness of our model across different generalization scenarios.}


\begin{document}

\section{Introduction}

The accurate prediction of radio signal propagation is a fundamental problem in wireless communications, enabling efficient network planning, resource allocation, and performance optimization \cite{10443723}. The problem of pathloss map prediction is challenging both for outdoor \cite{popoola_outdoor_2018} and indoor \cite{electronics12030497} settings. Outdoor environments introduce inherent challenges due to their dynamic nature, weather-dependent factors, and diverse terrain variations, including buildings, trees, and other obstacles that cause diffraction and shadowing. Additionally, the presence of moving objects such as vehicles and pedestrians further complicates signal modeling by introducing time-varying interference and multipath effects. On the other hand, indoor environments present unique challenges for radio propagation modeling due to their complex geometries, diverse materials, and varying building layouts, which create unpredictable signal blockages and non-line-of-sight conditions. 

Traditional analytical models struggle to capture these intricate propagation effects, often requiring extensive parameter tuning and empirical measurements. {The study presented in} {\cite{nguyen2023deep}} {demonstrates the superior predictive capabilities of deep learning models in the context of pathloss estimation. Similarly,} Ref. {\cite{zhang2019path}} {conducts a comparative analysis between a traditional log-distance pathloss model, a shallow neural network, and statistical machine learning approaches, including support vector regression and random forests. The findings indicate that the classical physical model underperforms relative to learning-based methods across multiple evaluation metrics. While statistical models that approximate pathloss as a monotonic decay function of distance provide a coarse baseline, these estimations can be further refined through neural networks} {\cite{feng2025ipp}}{. Moreover, although ray-tracing simulations offer high accuracy, they are computationally intensive and less efficient than deep learning-based alternatives, as noted in} {\cite{pathlosstransformer}}. Data-driven machine learning (ML) approaches have emerged as a powerful alternative, offering the potential to learn complex patterns directly from data \cite{zhang2019path}. By leveraging large-scale datasets and advanced learning algorithms, these methods can adapt to diverse environments and provide more accurate pathloss predictions compared to conventional models.

A challenge was created \cite{outdoorchallenge} to test the generalizability of ML models for outdoor environments, where participants were tasked with predicting the pathloss value for each pixel of an outdoor environment given the city map and transmitter location. The authors of the same challenge later introduced the first indoor radio map prediction \mbox{challenge \cite{indoorchallenge}}. The objective of this new challenge was to predict the pathloss value for each pixel of a given building layout while incorporating key factors such as the reflectance and transmittance of walls and doors, the location of the transmitter, the carrier frequency, and the antenna radiation pattern. Unlike outdoor settings, where large-scale features dominate \cite{9488784}, indoor environments require fine-grained modeling due to the intricate interactions between radio waves and surrounding structures.

Following our previous works on using such approaches for closely related problems, such as wireless positioning \cite{KHACHATRIAN2025103696} and environment map reconstruction \cite{reconstructioniwcmc}, we shift our methodological design from convolutional neural networks \cite{9451544} to vision transformers \cite{10.1145/3505244}. This shift is motivated by the superior capability of vision transformers in capturing long-range dependencies and global context, which are crucial for accurate pathloss predictions in complex indoor settings. {Moreover, transformers have been proven to perform better, as the dataset size and model capacity increases} {\cite{kaplan2020scaling,zhai2022scaling}}. However, unlike our previous works, where the primary focus was on large-scale spatial relationships, the model needs to be better suited for indoor environments, where localized interactions and fine structural details play a more significant role. {To address the limitations imposed by the small dataset size, we employ a diverse set of data augmentation techniques, including flipping, rotation, and MixUp. These augmentations help mitigate overfitting and contribute to improved generalization across unseen spatial configurations, frequency bands, and antenna types.}

\section{Related Works}

Numerous methodologies have been proposed to address the challenge of pathloss prediction in both outdoor and indoor environments \cite{6165627,oladimeji_propagation_2022}. Convolutional encoder--decoder architectures have demonstrated strong performance in this domain, effectively capturing spatial dependencies for accurate pathloss estimation \cite{qiu2023deep,lee2024scalable}. 

{The study presented in} {\cite{pathlossunet}} {introduces a UNet-based architecture designed for efficient radio map prediction in urban environments with mobile base stations and user equipment. Another method proposed in} {\cite{qiu2023deep}} {utilizes a SegNet-based framework to address the challenges associated with outdoor pathloss radio map estimation. In} {\cite{outdoorpathloss2}}, {a line-of-sight (LoS) map is generated and incorporated as an input feature to the neural network: a feature engineering strategy particularly beneficial in scenarios where data availability is limited.}

A {UNet-based} architectural framework, combined with a carefully designed loss function, enabled the authors of \cite{lu2025sip2net} to achieve first place in the first indoor pathloss radio map prediction challenge \cite{indoorchallenge}. Additionally, meticulous feature engineering—particularly incorporating the number of walls between each point within a building and the transmitter—has been shown to significantly enhance indoor pathloss prediction accuracy \cite{feng2025ipp}. {An alternative and noteworthy method, TransPathNet, proposed in} {\cite{indoorpathloss4}}, {tackles the problem of indoor radio map prediction through a two-stage pipeline: the first stage generates a coarse estimation of the pathloss distribution, which is subsequently refined by a second network to enhance the spatial and signal detail fidelity.}

{However, convolutional neural networks require deeper architectures to expand their effective receptive field, thereby enabling the modeling of long-range dependencies in spatial environments. In contrast, transformers are inherently capable of capturing such dependencies through self-attention mechanisms. The work by} {\cite{pathlosstransformer}} {demonstrates the successful application of transformer architectures to the task of radio map prediction, highlighting improvements in both predictive accuracy and computational efficiency. Moreover, empirical studies have demonstrated that increasing both dataset size and model capacity leads to substantial performance gains in transformer-based models} {\cite{kaplan2020scaling, zhai2022scaling}.}

{Inspired by these works}, our method leverages pretrained vision transformers to address the indoor pathloss prediction problem. Vision transformers, known for their ability to model long-range dependencies and capture complex spatial relationships, offer a promising alternative to convolutional architectures in this context. To address the data scarcity problem, data augmentation techniques are employed to enhance model generalization. Specifically, we utilize transformations such as flipping, rotation, random cropping, and MixUp to introduce variability and improve robustness. Our work is a natural continuation of our participation in the first indoor pathloss radio map prediction challenge \cite{ourindoor}.

\section{Data Description}
\label{sec:data}
The dataset utilized in this study comprises pathloss radio maps generated through a ray-tracing {algorithm}, which simulates indoor wireless signal propagation under diverse environmental conditions. These maps serve as a benchmark for assessing the generalization capabilities of neural networks in indoor radio map reconstruction tasks.

\subsection{Dataset Overview}

The dataset includes pathloss radio maps corresponding to 25 distinct building layouts, each characterized by specific wall placements as well as their respective transmittance and reflectance properties. Signal propagation is simulated at three carrier frequencies—\mbox{868 MHz}, 1800 MHz, and 3500 MHz—capturing the frequency-dependent effects of indoor environments. Additionally, five different antenna configurations are incorporated, each defined by its unique radiation pattern, with the first configuration corresponding to an isotropic antenna. In every scenario, the transmitting antenna (Tx) is positioned at a height of 1.5 m above the floor.

\begin{figure}[H]
{\captionsetup{position=bottom,justification=centering}
    \subfloat[Reflectance\label{fig:reflectance}]{
        \includegraphics[width=0.3\textwidth]{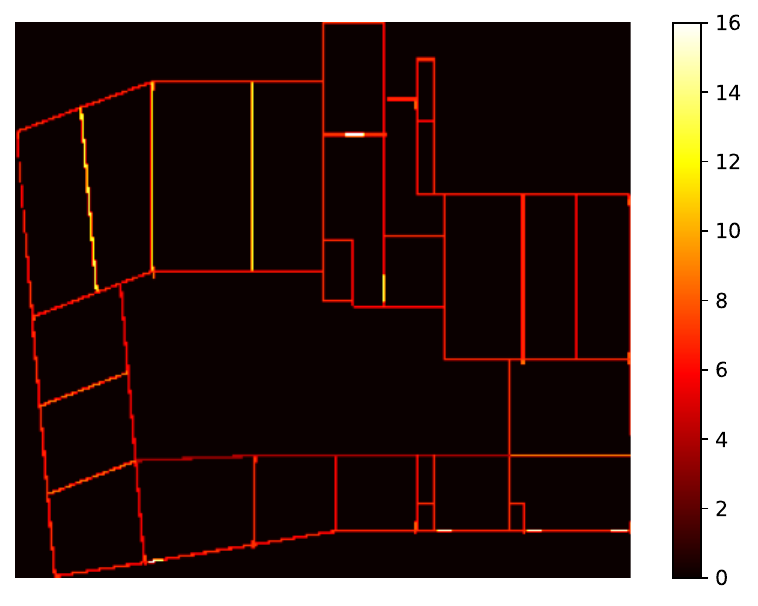}
    }\hfill
    \subfloat[Transmittance\label{fig:transmittance}]{
        \includegraphics[width=0.3\textwidth]{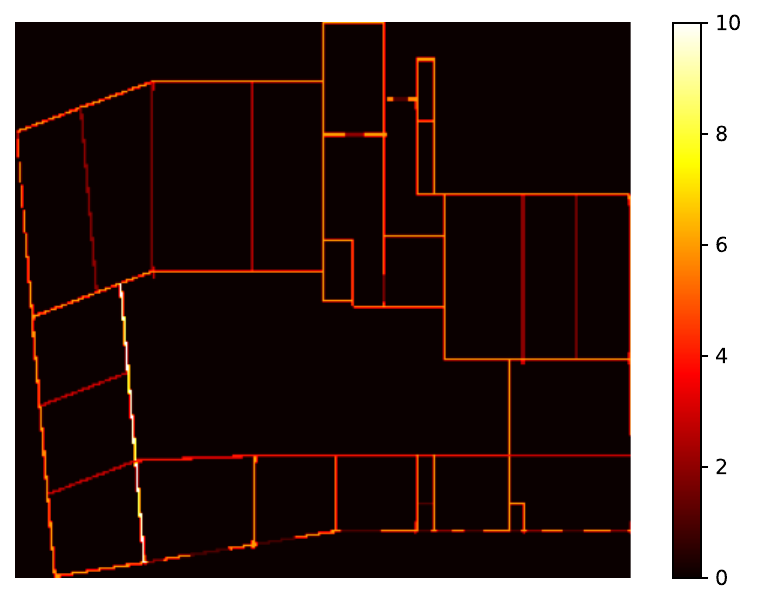}
    }\hfill
    {\captionsetup{position=bottom,justification=centering}}
    \subfloat[Distance from antenna\label{fig:distance}]{
        \includegraphics[width=0.3\textwidth]{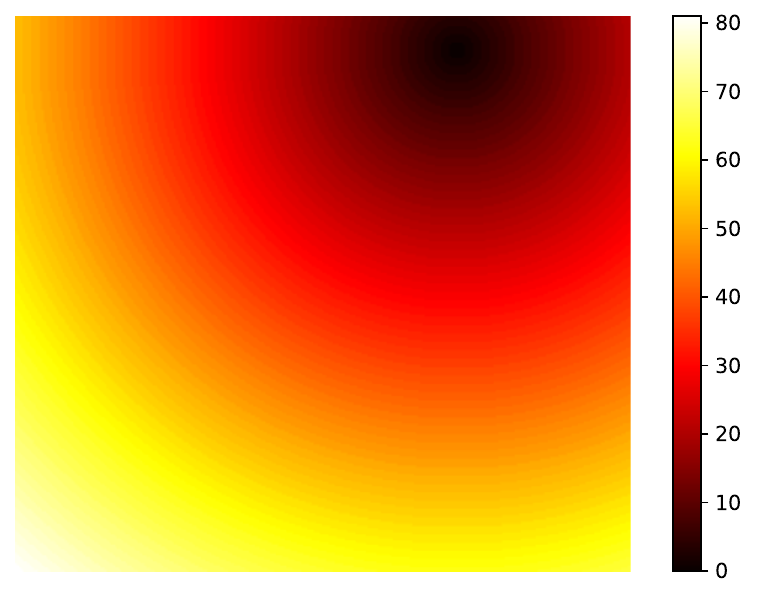}
    }}
{\captionsetup{position=bottom,justification=centering}
    \subfloat[Radiation pattern\label{fig:rp}]{
        \includegraphics[width=0.3\textwidth]{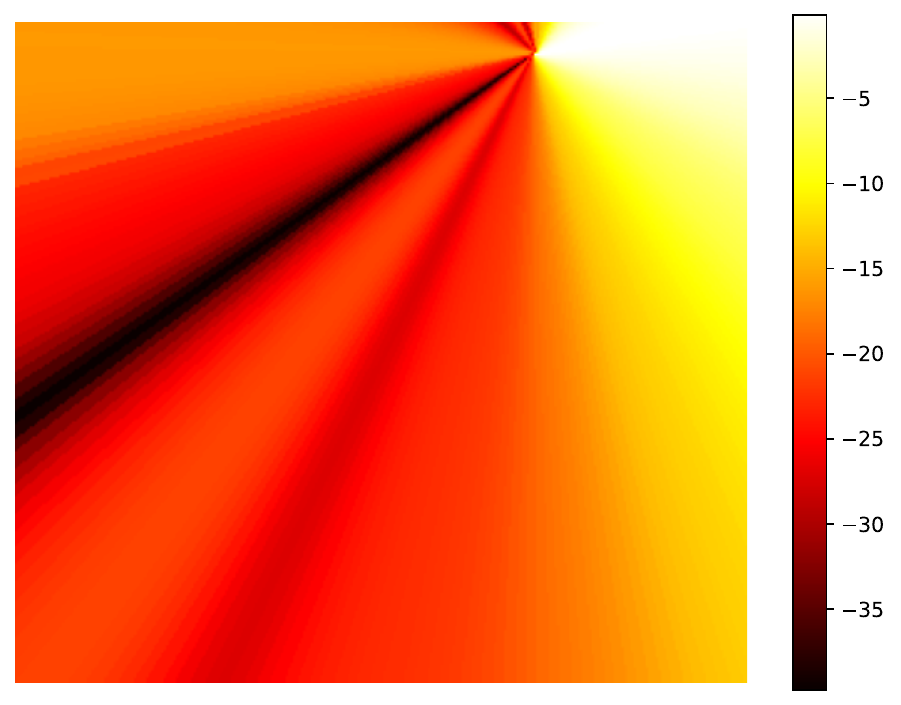}
    }
    \subfloat[Target\label{fig:target}]{
        \includegraphics[width=0.3\textwidth]{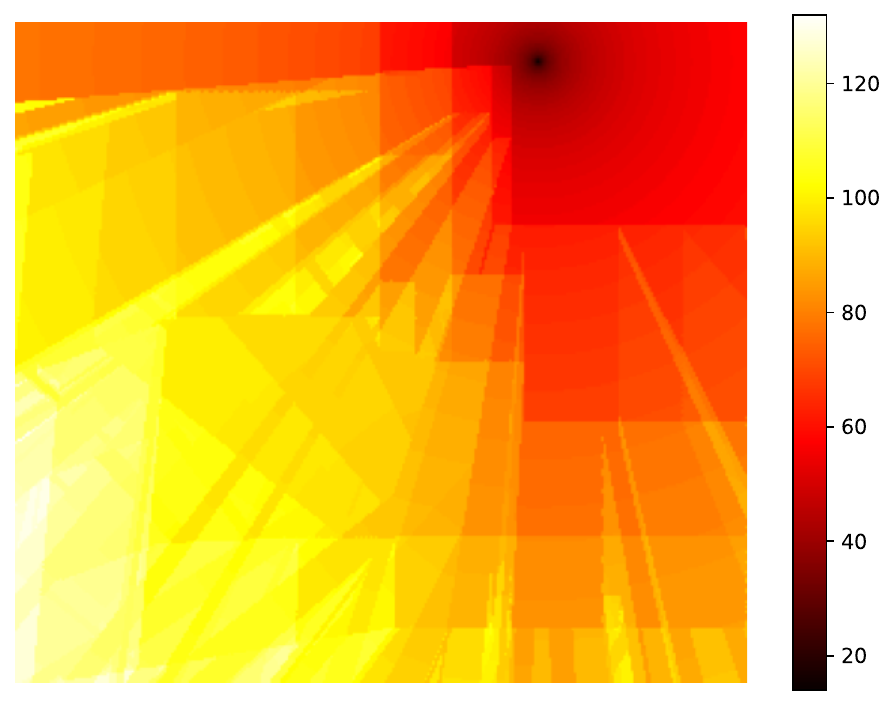}
    }}

    \caption{An example of the three given input channels (\textbf{a}--\textbf{c}), the radiation pattern channel that we created (\textbf{d}), and the target (\textbf{e}).}
    \label{fig:input-target}
\end{figure}

The buildings and antenna locations are encoded in three-channel images. Each pixel corresponds to a grid cell of size $0.25 \times 0.25$ m. The first channel contains absolute values for normal incidence reflectance of the walls ($0$ for air, Figure \ref{fig:input-target}a). The second channel contains absolute values for normal incidence transmittance of the walls ($0$ for air, \mbox{Figure \ref{fig:input-target}b}). The third channel encodes the distance from the antenna in meters (Figure \ref{fig:input-target}c).

\subsection{Task Descriptions}

To evaluate the generalization capabilities of the models under increasingly challenging conditions, three tasks are defined:

\begin{itemize}
    \item Task 1: This task assesses the model's ability to generalize across different building layouts. The evaluation is conducted using an isotropic antenna operating at a carrier frequency of 868 MHz, ensuring that variations in the test set arise only from differences in building structures.
    
    \item Task 2: \textls[-15]{In this setting, the model's generalization is evaluated across both unseen building layouts and carrier frequencies. The training set includes data from isotropic antennas operating at 868 MHz, 1800 MHz, and 3500 MHz, requiring the model to learn frequency-dependent propagation characteristics in addition to spatial variations.}  
    
    \item Task 3: The most challenging scenario is considered, where the model must generalize not only to unseen building layouts and carrier frequencies but also to unseen antenna configurations characterized by distinct radiation patterns. The training set comprises five different antenna configurations, introducing additional variability in wave propagation due to differences in antenna characteristics.  
\end{itemize}
\vspace{-12pt}

\subsection{The Challenge Test Set}

The challenge test set consists of 6 new building layouts. For Task 2, there are two carrier frequencies of 868 MHz and 2400 MHz. The test set contains three new radiation patterns for Task 3. A part of the test set, namely buildings \#1 and \#5, is available on Kaggle. Our main results are tested on the full challenge test set and evaluated by macro-averaged root-mean-square error (RMSE), while additional analysis are tested on the Kaggle test set and evaluated by micro-RMSE.

\section{Model Design and Neural Network Architecture}

Our approach employs deep learning techniques while relying on minimal prior knowledge of the underlying physics of radio signal propagation. Specifically, we utilize a pretrained neural network and fine-tune it on the available training data, leveraging extensive data augmentation strategies to enhance generalization and mitigate the limitations imposed by data scarcity.

\subsection{Data Preprocessing}

To ensure consistency and enhance feature representation, a series of systematic preprocessing steps were applied to the dataset. These steps standardize the input data, address variations in geometry and dimensions, and facilitate the integration of additional contextual information relevant to specific tasks. In alignment with the requirements of Tasks 1, 2, and 3, the dataset was prepared as follows:  

For Task 1, the input data consist of three channels: reflectance, transmittance, and distance. Each input image is padded to form a square and subsequently resized to \mbox{$518 \times 518$ pixels}. Padding is applied using a value of -1, as the value 0 holds meaningful significance within the dataset.  For Tasks 2 and 3, frequency information must also be incorporated. This is encoded as a single additional channel, where each pixel is assigned a uniform value corresponding to the carrier frequency in GHz. In Task 3, antenna configuration must be explicitly represented. To encode the antenna's radiation pattern, an additional input channel of the same spatial dimensions is introduced. Each pixel value corresponds to the antenna gain at the angle between the respective pixel position and the antenna location. This representation provides a structured visualization of the antenna's signal coverage, offering spatial context for signal propagation and station placement, as illustrated in Figure \ref{fig:input-target}d.

\subsubsection{Normalization}
We conducted an exploratory analysis of the provided dataset to determine the value ranges for each input channel. Based on these observations, we selected channel-specific normalization factors: 25 for reflectance, 20 for transmittance, 200 for distance, 10 for frequency, and 40 for the radiation pattern. Each channel was subsequently normalized by dividing its values by the corresponding factor. The resulting normalized channels serve as the input for the models.

\subsubsection{Data Augmentation}
We applied a series of data augmentation techniques to enhance the model’s generalization capabilities and robustness to variations in the input data. The following augmentation strategies were employed:  

\begin{itemize}  
    \item MixUp augmentation: MixUp is a data augmentation technique that generates synthetic training samples by linearly interpolating pairs of input images and their corresponding labels. This method enhances the model’s ability to generalize by encouraging smoother decision boundaries. Additionally, MixUp blends the frequency channels of two inputs, effectively creating intermediate frequency values and enabling the model to generalize to unseen frequencies. In our training pipeline, MixUp was applied to \(75\%\) of the training samples, while the remaining \(25\%\) were left unchanged. This balanced approach ensures that the model benefits from both augmented and original data, promoting diversity in the learned feature representations while preserving alignment with the original data distribution.  

    \item Rotation and flipping: These augmentations introduce variations in the spatial orientation of input samples, improving the model's invariance to geometric transformations. During training, each input sample is randomly rotated by \(0^\circ\), \(90^\circ\), \(180^\circ\), or \(270^\circ\), with equal probability. Since no transformation occurs when \(0^\circ\) is selected, rotation is effectively applied to approximately \(75\%\) of the training data. Similarly, flipping is applied, where each input sample, along with its corresponding label, is either left unchanged or flipped horizontally and/or vertically. These transformations mitigate overfitting to specific orientations and structural layouts, allowing the model to learn orientation-invariant representations and improve its generalization to unseen~data.  

    \item Cropping and resizing: Cropping is used to simulate partial observations of the input data, forcing the model to learn from varying spatial contexts. This augmentation is applied to \(75\%\) of the training samples, while the remaining \(25\%\) remain unaltered. The crop size is randomly selected within a range between half the size of the input image (\(259 \times 259\) pixels) and the full size (\(518 \times 518\) pixels). Following cropping, each extracted region is resized to \(518 \times 518\) pixels to maintain consistency in the input dimensions required by the model.  
\end{itemize}

\begin{figure}[H]
{\captionsetup{position=bottom,justification=centering}
    \subfloat[Reflectance\label{fig:reflectance-mixup}]{
        \includegraphics[width=0.32\textwidth]{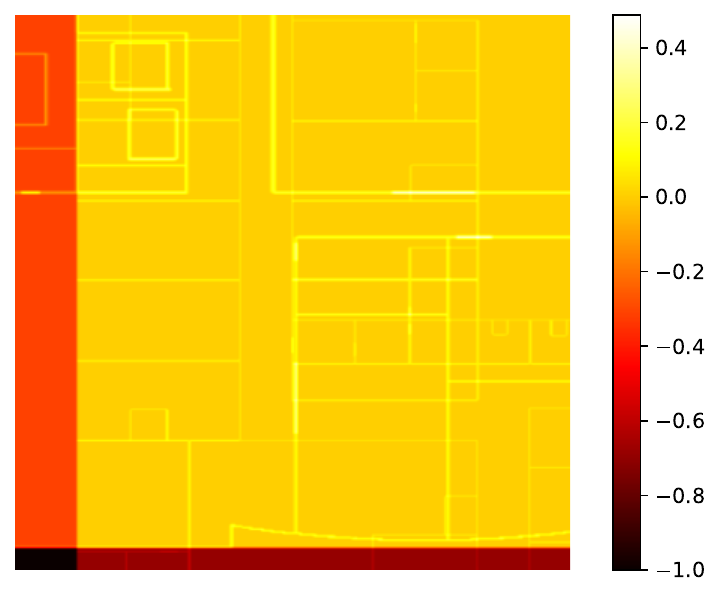}
    }\hfill
    \subfloat[Transmittance\label{fig:transmittance-mixup}]{
        \includegraphics[width=0.32\textwidth]{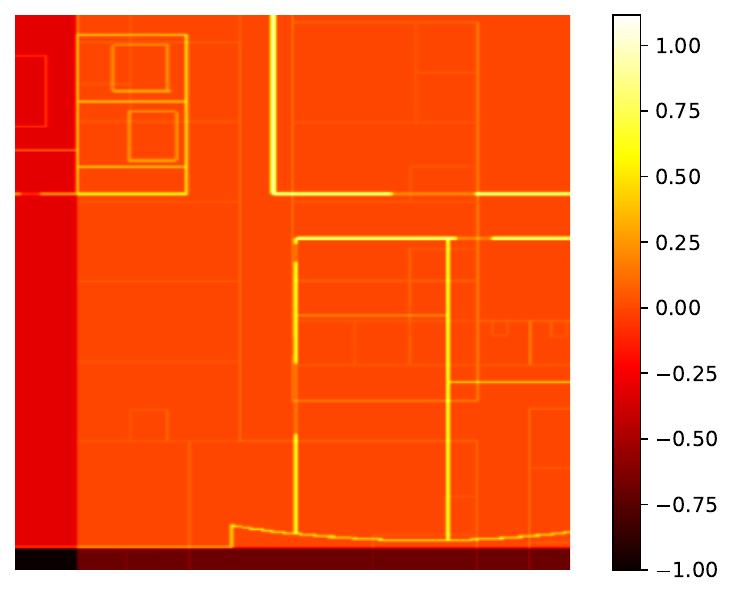}
    }\hfill
    \subfloat[Distance from antenna\label{fig:distance-mixup}]{
        \includegraphics[width=0.32\textwidth]{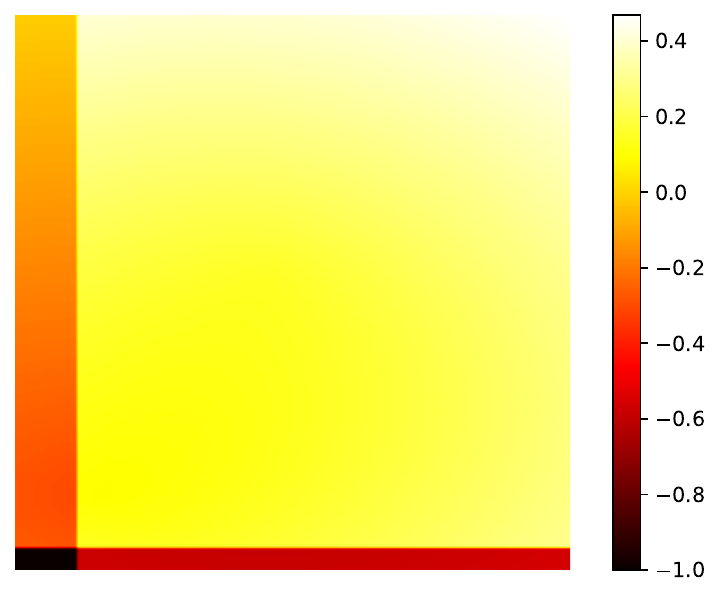}
    }

    \subfloat[Radiation pattern\label{fig:rp-mixup}]{
        \includegraphics[width=0.32\textwidth]{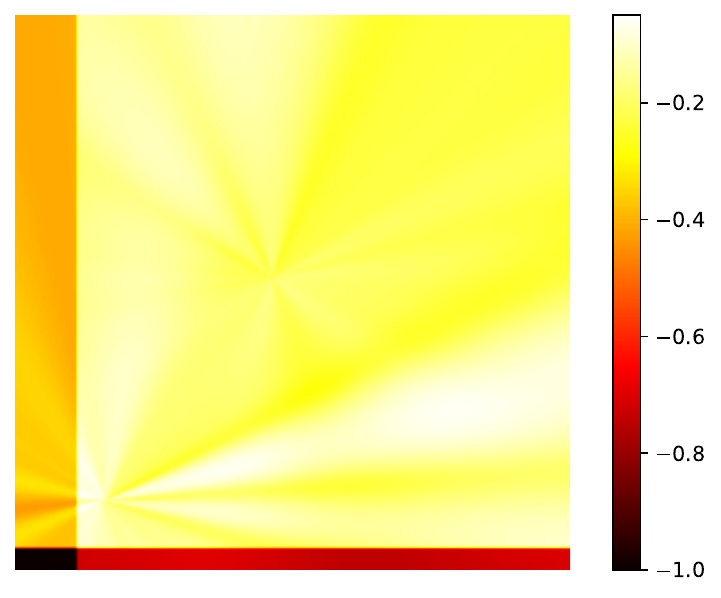}
    }
    \subfloat[Target\label{fig:target-mixup}]{
        \includegraphics[width=0.32\textwidth]{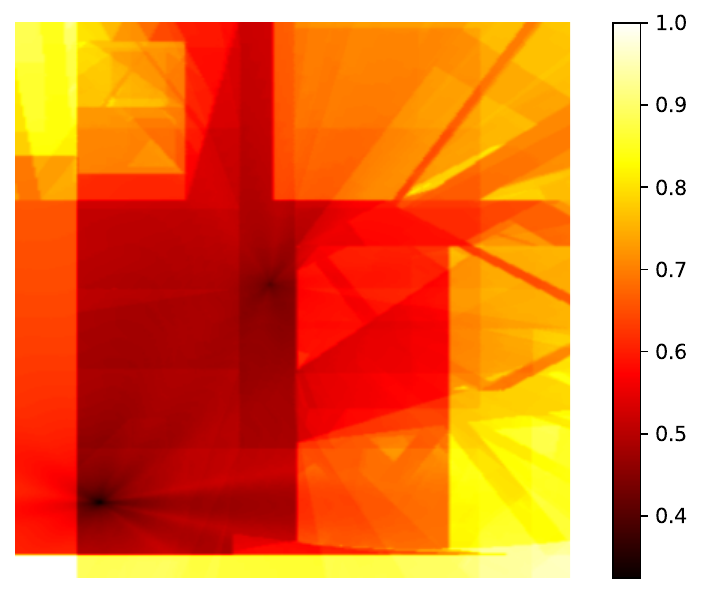}
    }}

    \caption{An example of the  input (\textbf{a}--\textbf{d}) channels, and the target (\textbf{e}) for a training sample after applying all the augmentations, normalization, and padding.}
    \label{fig:mixup}
\end{figure}

Figure~\ref{fig:mixup} depicts the obtained inputs after applying the aforementioned augmentations.

\subsubsection{Training, Validation, and Testing Splits}
For our experiments, the dataset was partitioned into training, validation, and testing sets according to predefined criteria. Across all tasks, the training set comprises \mbox{buildings \#1} to \#19, while buildings \#20 to \#22 are allocated to the validation set, and buildings \#23 to \#25 are reserved for testing.  For Tasks 2 and 3, the training set includes carrier frequencies of 868~MHz and 3500 MHz, whereas the 1800 MHz frequency is exclusively used for validation and testing.  For Task 3, the first three antenna configurations are included in the training set, while configuration \#4 is designated for validation, and configuration \#5 is held out for testing.  As a result, the validation and test sets contain previously unseen buildings across all tasks. Additionally, in Tasks 2 and 3, the validation and test sets include carrier frequencies not present in the training data. In Task 3, the model is further challenged by evaluating its performance on unseen antenna configurations in the validation and test~sets.

\subsection{Neural Network Design}

Our neural network consists of three parts: DINOv2 vision transformer \cite{dinov2}, used as an encoder, UPerNet convolutional decoder \cite{xiao2018unified}, and a neck responsible for connecting the ViT-based encoder and the convolutional decoder. We choose the ViT-B/14 version of DINOv2 with pretrained weights. First, the input image is passed through a convolutional layer, which outputs a three-channel image. We do this for compatibility with the DINOv2 input, to be able to leverage the pretrained weights of the network. The resulting image is passed to the encoder to obtain the embeddings of the image. Then, the embeddings from all the 14 layers (the first layer, 12 hidden layers, and the output layer) are passed through a linear layer to decrease the embedding size from 768 to 256 for task 1, and 512 for tasks 2 and 3. The low-dimensional embeddings are then reshaped into $\frac{518 \times 518}{14 \times 14} =37 \times 37$ squares and fed to the convolutional neck. 

{The neck consists of a convolutional layer with kernel size 1 that maps the neck input dimension to a predefined depth for each layer, followed up by a bilinear resize operation that scales the image by a predefined factor for each layer, and, finally, another convolutional layer with kernel size 3 that does not alter the depth of the input tensor. Earlier layers use a higher scale factor and smaller depth, while later layers are scaled with a smaller factor, and higher depth. Specifically, the scale factors for each layer are \{14, 14, 14, 8, 8, 8, 4, 4, 4, 2, 2, 2, 1, 1\}, and the convolution depths are \{16, 16, 16, 32, 32, 32, 64, 64, 64, 128, 128, 128, 256, 256\} for task 1, and \{32, 32, 32, 64, 64, 64, 128, 128, 128, 256, 256, 512, 512\} for task 2 and 3.}

\begin{figure}[H]
    \includegraphics[width=0.9\textwidth]{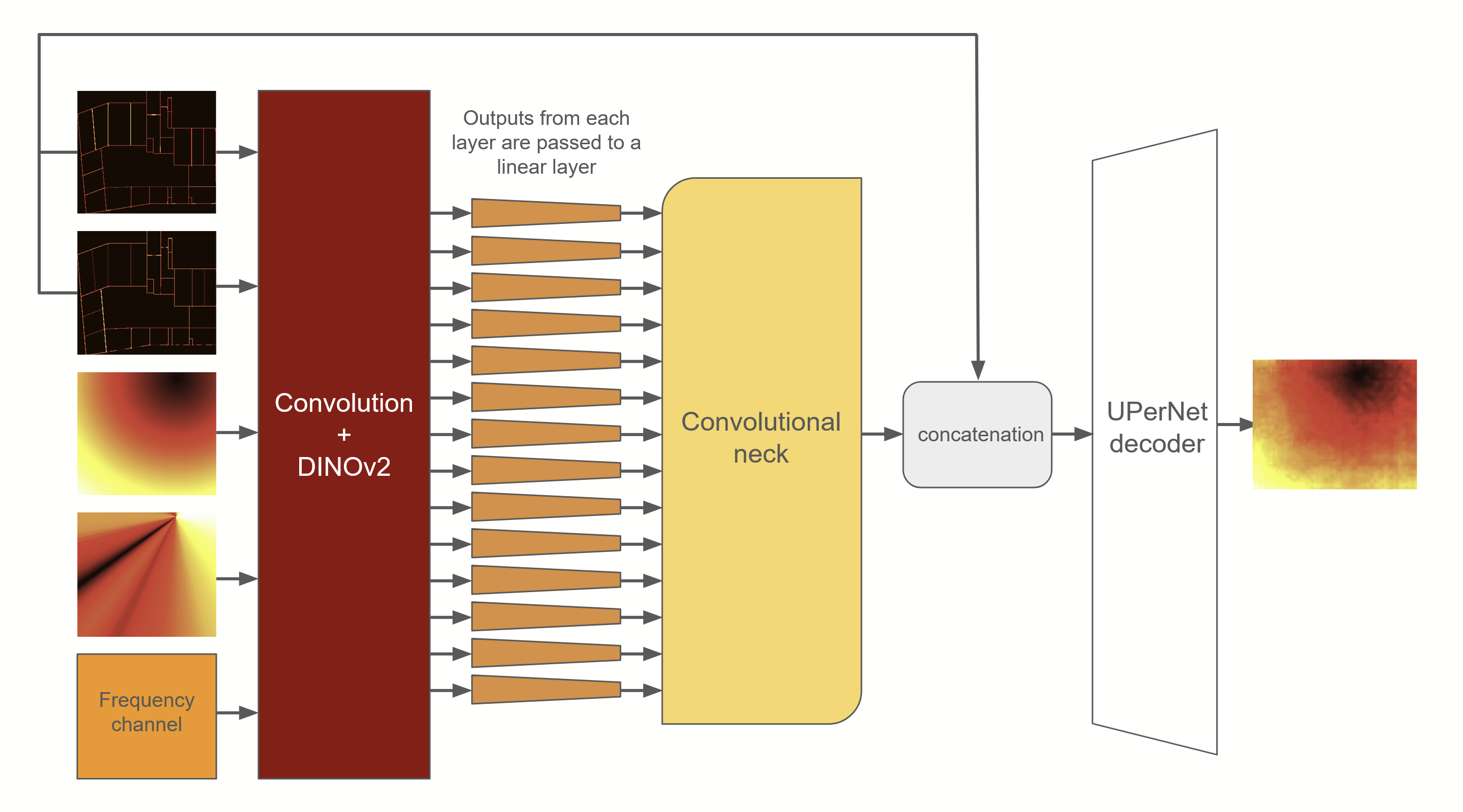}
    \caption{Model architecture.}
    \label{fig:model}
\end{figure}

The activations are then passed to the UPerNet {\cite{xiao2018unified}} decoder to obtain the output. To reinforce the room borders to the network, we also concatenate the reflectance and transmittance channels to the activations obtained from the neck, before feeding it to the decoder. The sigmoid function is then applied to the output, and the result is multiplied by 160 to obtain 
the final prediction. The model architecture can be seen in Figure~\ref{fig:model}.

\subsection{{Training Details}}

{The proposed network is trained using a mean squared error (MSE) loss function with a fixed learning rate of $3 \times 10^{-4}$. Training is conducted for 200 epochs on Task 1, 120 epochs on Task 2, and 40 epochs on Task 3. All experiments are performed on an NVIDIA DGX A100 system equipped with two 40 GB A100 GPUs, employing distributed data parallelism. The batch size is set to 10 for Task 1 and 4 for Tasks 2 and 3, corresponding to the maximum capacity permitted by GPU memory constraints. For all experiments, the DINOv2 module within our architecture is initialized with pretrained weights from the DINOv2-L/14 vision transformer. Post-training, the model checkpoint with the lowest validation loss is selected for evaluation. To ensure reproducibility, we utilize the \texttt{seed\_everything} function from the PyTorch 2.1.2 Lightning framework with a fixed seed value of 0.}

\section{{Experiments}}

{In this section, we present a series of experiments on Task 1 aimed at identifying the optimal model configuration. Specifically, we perform hyperparameter tuning to determine the most effective neural network settings, evaluate various data augmentation strategies to identify the most beneficial combination, and empirically demonstrate the contribution of manually crafted features under limited data conditions.}

\subsection{Optimal Architecture}

The architecture described in the previous section was selected based on multiple experimental decisions, all evaluated on Task 1 using our test set. As shown in Table~\ref{tab:architecture-results}, enforcing wall information through a skip connection led to a slight improvement in model performance. Conversely, increasing the size of the neck resulted in a deterioration in the RMSE score, as evidenced by the performance drop between the second and third rows. We also experimented with masking loss values on padding pixels; however, this approach did not yield better results. Additionally, we conducted an experiment using a fixed pixel size by padding images to $550 \times 550$ without resizing. Only four images exceeded this size, which were first padded and then resized to match the fixed dimensions. During training, all images were randomly cropped to $518 \times 518$. Since this setting introduced a significant number of padding pixels, we performed the fixed-scale experiment exclusively with loss masking enabled. The results in Table~\ref{tab:architecture-results} indicate that this approach was not effective.

\begin{table}[H]
    \centering
    \caption{The macro-RMSE (dBm) for Task 1 measured on our test set for several experiments.}
    \label{tab:architecture-results}
   \begin{tabularx}{\textwidth}{CCCCC}
        \toprule
        \textbf{Neck Size} & \textbf{Enforcing Walls} & \textbf{Loss Masking} & \textbf{Fixed Scale} & \textbf{RMSE} \\
        \midrule
        256 & \xmark & \xmark & \xmark & 6.0 \\
        256 & \cmark & \xmark & \xmark & 5.9 \\
        512 & \cmark & \xmark & \xmark & 6.7 \\
        256 & \cmark & \cmark & \xmark & 6.8 \\
        256 & \cmark & \cmark & \cmark & 6.7 \\
        \bottomrule
    \end{tabularx}
\end{table}

\subsection{Optimal Augmentations}

To determine the optimal set of augmentations, we conducted a series of controlled experiments. The flipping augmentation was consistently applied, as it is a robust transformation that does not introduce interpolation artifacts. Similarly, rotating the input and the output by a multiple of $90^\circ$ was also performed across all the experiments. The results in Table~\ref{tab:augmentation-results} indicate that excluding the MixUp augmentation leads to improved performance on our internal test set but results in a decline in performance on the Kaggle subset. Consequently, we retained MixUp for subsequent experiments. Additionally, we evaluated a more aggressive rotation augmentation, where both the input and corresponding output are rotated by arbitrary angles. However, this approach yields suboptimal performance on both the internal and Kaggle test sets. Lastly, we implemented an improved interpolation technique in which each pathloss radio map $p$ is transformed into linear scale as $p' = 10^{\frac{p}{10}}$, followed by interpolation $I(p')$, and then converted back to decibels using $10 \times \log_{10} I(p')$. Surprisingly, this method also fails to enhance performance across both test sets.

\begin{table}[H]
    \centering
    \caption{The macro-RMSE (dBm) for Task 1 measured on our test set and Kaggle subset for different feature engineering methods.}
    \label{tab:augmentation-results}
    \begin{tabularx}{\textwidth}{CCCCC}
        \toprule
        \textbf{Arbitrary Rotations} & \textbf{MixUp} & \textbf{Proper Output Interpolation} & \textbf{Our Test} & \textbf{Kaggle Test} \\
        \midrule
        \xmark & \xmark & \xmark & 5.9 & 11.7 \\
        \xmark & 75\%   & \xmark & 6.2 & 8.7  \\
        75\%   & 75\%   & \xmark & 7.2 & 9.3  \\
        \xmark & 75\%   & \cmark & 6.7 & 11.0 \\
        \bottomrule
    \end{tabularx}
\end{table}

\subsection{Feature Engineering}

Building on the insights from \cite{feng2025ipp}, we propose three feature engineering techniques and assess their impact on both our internal and Kaggle test sets. The first approach introduces an additional input channel that encodes free-space pathloss (FSPL) information. The second method computes the number of obstructions between each pixel and the transmitter. The third approach combines both features by taking their sum. The results of these experiments are presented in Table~\ref{tab:features-results}. Our findings indicate a significant performance improvement when incorporating the obstruction feature. However, providing FSPL estimates as an additional input negatively affects model performance.

\begin{table}[H]
    \centering
    \caption{The macro-RMSE (dBm) for Task 1 measured on our test set and Kaggle subset for several augmentation combinations.}
    \label{tab:features-results}
 \begin{tabularx}{\textwidth}{CCCC}
        \toprule
        \textbf{FSPL} & \textbf{Obstructions} & \textbf{Our Test} & \textbf{Kaggle Test} \\
        \midrule
        \xmark & \xmark & 6.2 & 8.7  \\
        \cmark & \xmark & 6.7 & 11.4 \\
        \xmark & \cmark & 4.3 & 7.2  \\
        \cmark & \cmark & 4.7 & 7.7  \\
        \bottomrule
    \end{tabularx}
\end{table}

\section{Results}

Table~\ref{tab:results} presents a summary of the performance across all three tasks. We report results on both the {Kaggle} test set and the official test set from the first indoor pathloss radio map prediction challenge \cite{indoorchallenge}, where our approach achieved eighth place. The findings indicate that our training strategy demonstrates strong robustness to variations in frequencies and antenna types. Notably, the performance drop between Task 1 and the other two tasks remains within 3.5 absolute points, or 31\% in relative terms, on the challenge test set. This degradation is lower than what we observed on our own test set and is comparable to the top-performing solutions in the competition. {After selecting the best combination of neural network hyperparameters, augmentations, and manually engineered features, our best models outperforms our initial submission across all three tasks, both on the Kaggle subset and the entire challenge test set.}

{Additionally, we compare our model to the runner-up and the winner of the challenge.}

\begin{table}[H]
    \centering
    \caption{The RMSE (dBm) measured on the Kaggle test set and the private challenge test set for the three tasks.}
    \label{tab:results}
     \begin{tabularx}{\textwidth}{CCCCCC}
        \toprule
        \shortstack{\textbf{{Evaluation}} \\ \textbf{{Set}}} & \textbf{Task} & \shortstack{\textbf{Our} \\ \textbf{Submission}} & \shortstack{\textbf{{Our Best}} \\ \textbf{{Model}}} & \textbf{{Runner-Up}} & \textbf{Winner} \\
        \midrule
        \multirow{4}{*}{{Kaggle}} 
            & {Task 1} & {8.7}  & {7.2}  & {5.3}  & {5.3}  \\
            & {Task 2} & {16.7} & {14.7} & {10.6} & {10.5} \\
            & {Task 3} & {17.4} & {15.7} & {12.6} & {12.0} \\
            & {Weighted Average} 
                          & {14.6} & {12.9} & {9.8}  & {9.5}  \\
        \midrule
        \multirow{4}{*}{Challenge} 
            & Task 1 & 11.3 & {11.0} & {7.8}  & 7.9 \\
            & Task 2 & 14.8 & {13.2} & {10.2} & 10.1 \\
            & Task 3 & 14.7 & {13.4} & {10.3} & 10.1 \\
            & Weighted Average 
                     & 13.7 & {12.6} & {9.5} & 9.4 \\
        \bottomrule
    \end{tabularx}
\end{table}

\subsection*{Detailed Evaluation}

Our test split is designed to isolate and analyze the individual effects of each axis of generalization for Tasks 2 and 3. Specifically, we evaluate our models on all possible combinations of seen and unseen buildings, frequencies, and, in the case of Task 3, antennas. As shown in Table~\ref{tab:detailed-eval}, our model for Task 2 exhibits an unexpected improvement in performance on unseen frequencies compared to seen ones, a trend that is also observed for Task 3. Additionally, our findings indicate that the most challenging axis of generalization for our models is adapting to previously unseen buildings.

\begin{table}[H]
    \centering
    \caption{Detailed evaluation for Tasks 2 and 3.}
   \begin{tabularx}{\textwidth}{m{7cm}LCC}
        \toprule
        \textbf{Test Set} & \textbf{Task 2} & \textbf{Task 3} \\
        \midrule
        Unseen Building  Frequency Antenna & 10.8 & 12.1  \\
        Unseen Building Frequency Antenna & 8.0  & 9.8   \\
        Unseen Building Frequency Antenna  & -    & 10.1  \\
        Unseen Building Frequency Antenna  & 9.1  & 11.0  \\
        Unseen Building Frequency Antenna  & -    & 13.1  \\
        Unseen Building Frequency Antenna  & -    & 10.2  \\
       Unseen Building Frequency Antenna  & -    & 11.7 \\
        \bottomrule
    \end{tabularx}
    \label{tab:detailed-eval}
\end{table}

\section{On Distribution Shift}

\begin{figure}[H]
{\captionsetup{position=bottom,justification=centering}
    \subfloat[Challenge test set\label{fig:sample-kaggle}]{
        \includegraphics[width=0.55\textwidth]{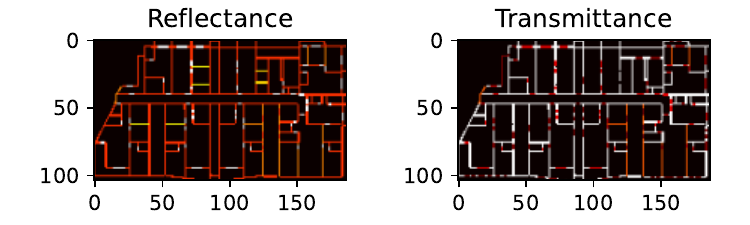}
    }\\
    \subfloat[Our test set\label{fig:sample-our-test}]{
        \includegraphics[width=0.55\textwidth]{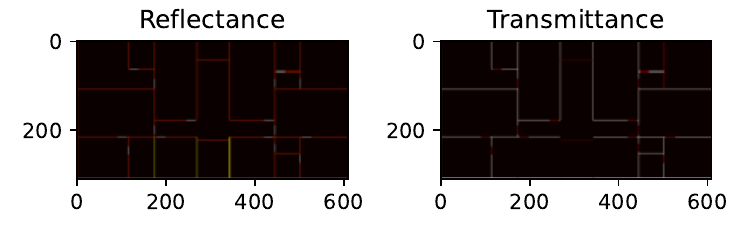}
    }\\
    \subfloat[Manually generated crop\label{fig:sample-crop}]{
        \includegraphics[width=0.55\textwidth]{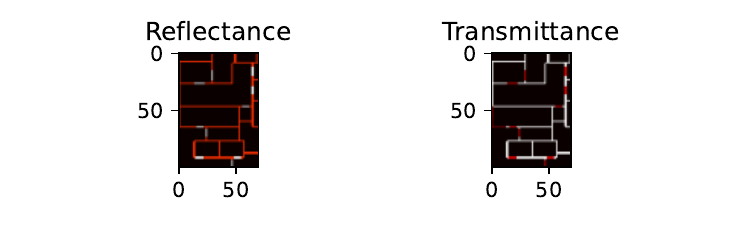}
    }}
    \caption{Input examples from the challenge test set (\textbf{a}), from our test set (\textbf{b}), and from the crops that we generated (\textbf{c}).}
    \label{fig:dist-shift}
\end{figure}

\begin{figure}[H]
    \includegraphics[width=1\linewidth]{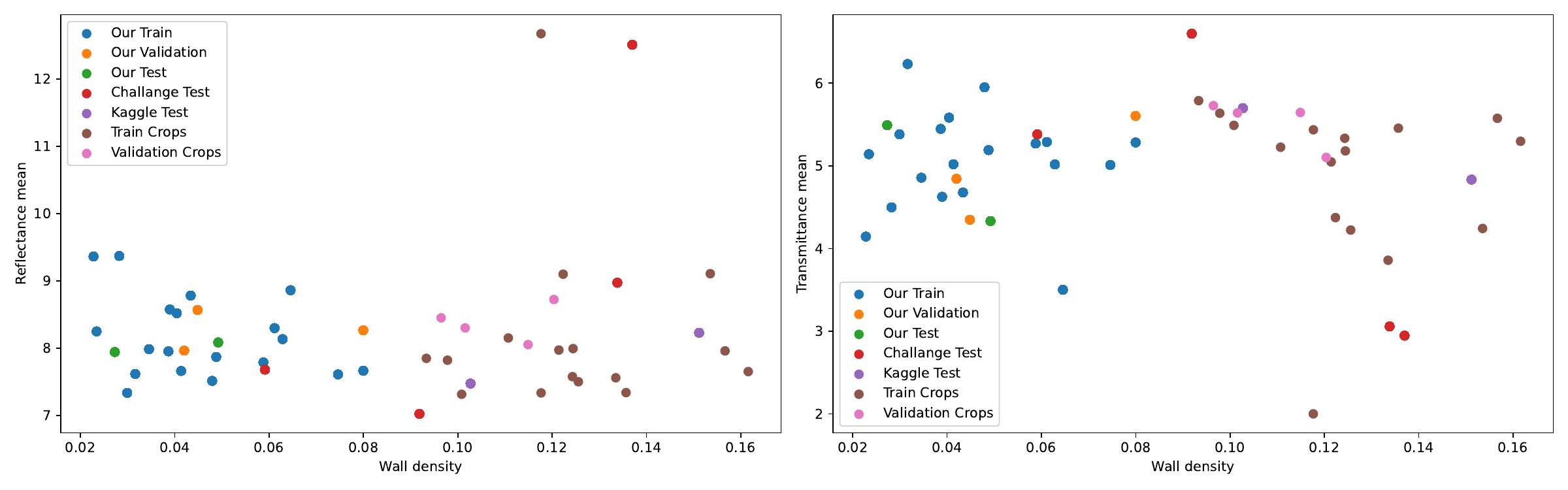}
    \caption{Wall density vs the average value of transmittance and reflectance on the walls. The figure indicates the distribution shift from our dataset to the challenge test set and how we tried to address it with manual crops.}
    \label{fig:wall-density}
\end{figure}

Our analysis revealed that the substantial discrepancy between our test set and the challenge test set primarily arises from a distribution shift in building layouts. The challenge test set consists of significantly smaller layouts with a higher density of walls. Furthermore, data points from the challenge test set exhibit increased reflectance and transmittance values compared to our test set. To mitigate this distribution shift, we manually selected buildings from our training and validation sets and extracted regions characterized by dense wall structures with high reflectance and transmittance values. These cropped sections were then incorporated into our training and validation sets. Figure~\ref{fig:dist-shift} illustrates random samples from the challenge test set, our test set, and the manually generated crops. To quantitatively assess the quality of our cropped regions, we compute the proportion of wall pixels relative to the total number of pixels, as well as the average transmittance and reflectance values, excluding air. Figure~\ref{fig:wall-density} presents the relationship between wall density and the average transmittance and reflectance. These values are calculated across our training, validation, and test sets, as well as the challenge test set, its Kaggle subset, and the manually extracted crops from our training and validation sets. The visualization in Figure~\ref{fig:wall-density} provides insight into the extent of the distribution shift and illustrates the extent to which the manual crops contributed to mitigating it.

However, as shown in Table~\ref{tab:dist-shift-results}, this approach was insufficient to fully capture the underlying characteristics of the challenge test set. Our results suggest that the pathloss values of the cropped sections are influenced by other parts of the building that were not included in the extracted regions, highlighting the complexity of the distribution shift.

\begin{table}[H]
    \centering
    \caption{The micro-RMSE (dBm) measured on our test set and the Kaggle test set for the three tasks.}
    \label{tab:dist-shift-results}
   \begin{tabularx}{\textwidth}{CCCC}
        \toprule
        \textbf{Crops} &  \textbf{Task} & \textbf{Our Test} & \textbf{Kaggle Test} \\
        \midrule
        \xmark & Task 1 & 6.2  & 8.7  \\
        \cmark & Task 1 & 6.8  & 8.3  \\
        \xmark & Task 2 & 9.3  & 16.7 \\
        \cmark & Task 2 & 9.3  & 18.1 \\
        \xmark & Task 3 & 11.5 & 17.4 \\
        \cmark & Task 3 & 11.8 & 18.5 \\
        \bottomrule
    \end{tabularx}
\end{table}

\section{Conclusions}

In this work, we tackled the complex problem of indoor pathloss prediction using a ViT-based architecture incorporating DINO-v2 pretrained weights. We systematically explored various architectural choices through extensive experimentation to determine the most effective model configuration. Our empirical findings demonstrate that extensive data augmentation plays a crucial role in mitigating overfitting, thereby improving model generalization. Furthermore, we highlight the significance of feature engineering in addressing challenges arising from data scarcity. A detailed evaluation was conducted to assess model performance across different axes of generalization. Additionally, we attempted to mitigate distribution shift by manually cropping regions of existing maps. However, this approach proved ineffective, as it inherently assumes that the cropped areas are independent of the excluded regions, which does not hold in practice. {We also compared our model to state-of-the-art approaches. The novelty of each approach is summarized in} {Table~\ref{tab:method-innovations}.}

\begin{table}[H]
\caption{{Comparison of methods and their key innovations.}}
\begin{tabularx}{\textwidth}{Lm{10cm}L}
    \toprule
    {\textbf{Method}} & {\textbf{Key Innovations}} \\
    \midrule
    {SIP2Net (Winner)} & {Incorporates a custom loss function and employs a discriminator network to distinguish predicted from real pathloss maps.} \\
    \midrule
    {IPP-Net (Runner-up)} & {Utilizes curriculum learning and introduces a wall-counting feature that encodes the number of obstructions from the transmitter.} \\
    \midrule
    {Ours} & {Leverages vision transformers and applies extensive data augmentation to enhance model generalization.} \\
    \bottomrule
\end{tabularx}

\label{tab:method-innovations}
\end{table}

\section{{Future Directions}}

{A promising direction for future work involves evaluating the scalability of our proposed method within the domain of pathloss radio map prediction, particularly through the generation of large-scale synthetic datasets. Additionally, developing more effective strategies to mitigate distribution shift remains an open and challenging research problem. Another interesting avenue is the integration of sparse real-world measurements to enhance the accuracy of predicted radio maps.}

\vspace{6pt} 
\authorcontributions{Conceptualization, H.K. and T.P.R.; Methodology, R.M. and H.K.; Software, R.M., E.G. and K.P.; Validation, R.M., E.G. and K.P.; Investigation, H.K. and T.P.R.; Data curation, R.M.; Writing---original draft, R.M.; Writing---review \& editing, H.K. and T.P.R.; Visualization, R.M.; Supervision, H.K. and T.P.R.; Project administration, H.K. and T.P.R. All authors have read and agreed to the published version of the manuscript. }

\funding{The work of R.~Mkrtchyan, E.~Ghukasyan, and H.~Khachatrian was partly supported by the RA Science Committee grant No. 22rl-052 (DISTAL). The work of T.~Raptis was partly supported by the European Union, Next Generation EU, under the Italian National Recovery and Resilience Plan (NRRP), Mission 4, Component 2, Investment 1.3, CUP B53C22003970001, partnership on ``Telecommunications of the Future'' (PE00000001---program ``RESTART'').
}

\dataavailability{Data are contained within the repository \cite{c0ec-cw74-24}. 
}

\conflictsofinterest{The authors declare no conflicts of interest.}

\begin{adjustwidth}{-\extralength}{0cm}

\reftitle{References}



\PublishersNote{}
\end{adjustwidth}
\end{document}